# Laplacian Mixture Model Point Based Registration

Mohammad Sadegh Majdi
Electrical Engineering Department.
Sharif University of technology
Tehran, Iran
msm2024@gmail.com

Emad Fatemizadeh
Electrical Engineering Department.
Sharif University of Technology
Tehran, Iran
fatemizadeh@sharif.edu

*Abstract*—Point base registration is an important part in many machine vision applications, medical diagnostics, agricultural studies etc. The goal of point set registration is to find correspondences between different data sets and estimate the appropriate transformation that can map one set to another. Here we introduce a novel method for matching of different data sets based on Laplacian distribution. We consider the alignment of two point sets as probability density estimation problem. By using maximum likelihood methods we try to fit the Laplacian mixture model (LMM) centroids (source point set) to the data point set.

*Keywords—registration , correspondence , alignment , rigid , laplacian , mixture model , affine ,point base*

## I. INTRODUCTION

Point based registration tries to find the best spatial transformation to fit one data set to another. The purpose of such action is to map different datasets into a globally consistent model and so be able to compare, extract features, and estimate their poses without the need of high level unnecessary mathematical dilemmas.

Point base registration is an important part in many image processing or machine vision based applications such as medical diagnostics from detection of cancer to study of changes in a tumor, angiography etc., agriculture for evaluating the process of products growing or the productivity of lands, military purposes in which it help us detect any suspicious movement from the foreign threats in our borders, the mentioned applications are just some of a very long list of registration application.

Point sets could refer to a vast variety of data from a 3D scanner output to any kind of feature extracted data.

## II. RELATED WORKS

The iterative closest point (ICP) algorithm was first introduced by Besl and McKay [19] it defined an iterative rigid algorithm. This method assumes that each point in source data sets A is corresponds to the closest point in the moving data set B and tries to minimum the square difference of two datasets. In EM-ICP [1] Antipolis and Neuilly assumed matches are hidden variable and tried to solve the equation using expectation maximization (EM) principles. Due to consideration of Gaussian noise, this method corresponds to a multiple matches ICP with normalized Gaussian weights. In [2] Andrew W. Fitzgibbon tried to minimize the registration error using nonlinear optimization (the Levenberg-Margquardt algorithm), they also introduce a data structure for minimization based on chamfer distance.

Robust point base registration has been proposed by Gold et al [3]. This method in contrast to ICP which used binary correspondences generated by nearest neighbor uses soft assignment of correspondences between two data sets.

Chui and Rangarajan [4] introduced the thin plate spline robust point matching (TPS-RPM) algorithm. By parameterization of transformation as a thin pate spline to this method tries to find the non-rigid registration to the RPM method.

Xiaoquang Hua et al [5] introduce an information geometry based algorithm for registration of point sets. By using Gaussian mixture model their method converts the Point sets to the statistical manifolds in which the dimension of statistical manifold is equal to the component of mixture model. It tries to find the shortest path between two manifold and then uses EM algorithm to solve the optimization problem.

Zhiyong Zhou et al introduced the Student's-t mixture model, a more robust method than Gaussian mixture model. One of the superiority of this method to the popular coherent points drift algorithm [6] is the less parameters it needs to adjust manually.

Onofrey et al. [7] presents a low-dimensional nonrigid registration method for fusing magnetic resonance imaging (MRI) and trans-rectal ultrasound (TRUS) in image-guided prostate biopsy. It develops a fast nonrigid registration on statistical deformation model.

T Economopoulos et al proposed an enhanced hexagonal center-based inner search (EHCBIS) algorithm, for automated point correspondence in dental image registration.

In this article we are proposing a new method based on popular coherent point drift [16] but instead of Gaussian mixture model we solved the problem using Laplacian mixture model [13, 14, 15].

## III. METHOD

### A. General methodology

The registration method can be summarized as follows, let's have two finite size point sets {A, B}, our goal is to find the best transformation model with parameters θ that map B to

A. in this article we suppose one data set represents the Laplacian mixture model (LMM) centroids and the second data set represents the constant data points. The optimum transformation will be calculated by maximizing the LMM posterior probability. Our method tries to preserve the topological structure of the data sets by forcing the LMM centroids to move coherently as a group.

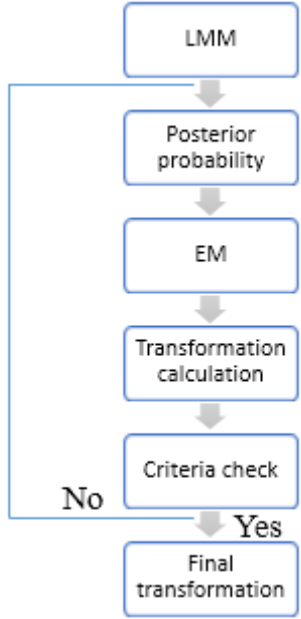

Fig. 1 Block diagram of proposed method

D indicates the dimension of point sets. N, M are number of points in data sets. The first point set $X_{N*D} = (x_1 \ldots x_N)^T$ is the source constant data set and the $Y_{M*D} = (y_1 \ldots y_M)^T$ is the LMM centroids or moving data sets.

We define the LMM probability density function as

$$\sum_{c=1}^{M+1} A(c) p(x|c), \quad (1)$$

$$p(x|c) = \frac{1}{2b} \exp\left(-\frac{|x-y_c|}{b}\right). \quad (2)$$

To estimate the unknown parameters we write a likelihood function as follows

$$L(b,s,R) = \sum_{n=1}^{N} Ln \sum_{m=1}^{M} A(c) p(x_n|c). \quad (3)$$

Since solving the prior probability of centroids is an ordeal, by using the Bayesian method we try to solve the posterior probability of the LMM centroids

$$P(c|x_n) = A(c) \frac{p(x_n|c)}{p(x_n)} \quad (4)$$

We also use the EM algorithm [9], [10] to find the unknown parameters θ. And so the final cost function will be

$$Q = -\sum_{n=1}^{N} \sum_{m=1}^{M} P^{old}(c|x_n) \log\left(A^{new}(c) p^{new}(x_n|c)\right), \quad (5)$$

Which $P^{old}$ is the values of Laplacian Distribution in last iteration and $P^{new}$ has the unknown parameters.

After substituting the equation (2) in equation (5) we'll have

$$Q = -\frac{1}{b} \sum_{n=1}^{N} \sum_{m=1}^{M} P^{old}(c|x_n) |x_n - T(y_c, \theta)|, \quad (6)$$

and

$$P^{old}(c|x_n) = \frac{\exp\left(-0.5 \left|\frac{x_n - T(\theta^{old}, y_c)}{b}\right|\right)}{\sum_{c=1}^{M} \exp\left(-0.5 \left|\frac{x_n - T(\theta^{old}, y_c)}{b}\right|\right)}. \quad (7)$$

But the problem here is that solving this equation for norm1 is very hard, so we should substitute the norm 1 with a norm 2 approximation.

There are different kinds of approximation for norm 0, so first we suppose norm 1 is almost equivalent to norm 0 and then substitute this norm 0 with an appropriate equivalent.

Some of equivalent estimations for norm 0 are

$$|x|_1 \approx |x|_0 \approx 1 - \exp(-x^2/\varepsilon^2) \approx \frac{x^2}{x^2 + \varepsilon^2} \approx \sum_{i=1}^{D} \omega_i x_i^2. \quad (8)$$

In this article we use the last estimation which changed the equation (6) to the following cost function.

$$Q = -\frac{1}{b} \sum_{n=1}^{N} \sum_{m=1}^{M} P^{old}(c|x_n) \sum_{i=1}^{2} \omega_i \left(x_n - T(y_c, \theta)\right)_i^2, \quad (9)$$

Where i stands for the dimension.

Now in order to find the parameters we just need to define a transformation function and solve the cost function.

We will solve this cost function for two common transformation, rigid and affine transforms.

B. Rigid Transform

We define the rigid transform as follows

$$T(\theta^{old}, y_c) = sRy_c + d, \quad (10)$$

Where s is the scale parameters to bring two datasets into a same size, this parameter will help us have a better measures of how close two data sets have gotten, R is a square D-dimensional matrix, and d is a D-dimensional vector which shows us the displacement required in each dimension.

So in order to solve this we start with displacement. By taking the partial derivative of Q with respect to d we will have

$$d = \frac{1}{N'} X^T P^T - \frac{1}{N'} sRY^T P = \mu_x - sR\mu_y, \quad (11)$$

And $N' = \sum_{n=1}^{N} \sum_{m=1}^{M} P^{old}(c | x_n)$.

Now by substituting equation (11) into (9) we'll have

$$Q = \frac{1}{b} \begin{bmatrix} tr\left(\sum_{i=1}^{D} \omega_i (A^T d(P^T 1) A)_i\right) - \\ 2str\left(\sum_{i=1}^{D} \omega_i (A^T P^T BR^T)_i\right) + \\ s^2 tr\left(\sum_{i=1}^{D} \omega_i (B^T d(P^T 1) B)_i\right) \end{bmatrix}, \quad \begin{matrix} A = X - \mu_x \\ B = Y - \mu_y \end{matrix} \quad (12)$$

So now we substitute the parts independent of Rotation with constant, the cost function will be

$$Q = \frac{1}{b}\left[-c_1 tr\left(\sum_{i=1}^{D} \omega_i ((A^T P^T B) R)_i\right) + c_2\right]. \quad (13)$$

For simplicity as we did in the displacement calculation, we try to take the derivative for each dimension separately, but in order to be able to do this for rotation, first we need to eliminate the sum equation, so we substitute it with a D-dimension diagonal matrix

$$\sum_{i=1}^{D} \omega_i \rightarrow W = \begin{bmatrix} \omega_1 & 0 & 0 \\ 0 & \ddots & 0 \\ 0 & 0 & \omega_D \end{bmatrix}, \quad (14)$$

By substituting W into the cost function it will be

$$Q = \frac{1}{b}\left[-c_1 tr(G^T R) + c_2\right], \\ G = W(A^T P^T B). \quad (15)$$

To solve this equation we need to use lemma 1 [11]

**Lemma 1.** If $R_{D*D}$ is an unknown rotation matrix and $G_{D*D}$ is an unknown real square matrix, let $USV^T$ be the singular value decomposition of G and

$$S = d(s_i), s_1 \geq s_2 \geq ... \geq s_D \geq 0,$$

Then the optimal rotation matrix R will be
$Q = UCV^T$ where $C = d(1,1,...,1,\det(UV^T))$

The $P^{old}$ is

$$P^{old}(c | x_n) = \frac{\exp\left(-0.5 \left|\frac{x_n - s^{old} R^{old} y_c - t^{old}}{b}\right|\right)}{\sum_{c=1}^{M} \exp\left(-0.5 \left|\frac{x_n - s^{old} R^{old} y_c - t^{old}}{b}\right|\right)}. \quad (16)$$

*C. Affine transform*

Affine transform can be defined as
$T(\theta^{old}, y_c) = By_c + d$ which $B_{D*D}$ is an affine transformation matrix, $d_{D\times 1}$ is the displacement vector, counting for the displacement in each dimension.

The unknown parameters for affine transformation will be as follows

$$d = \mu_x - B\mu_y. \quad (17)$$

$$B = (A^T P^T B) * (B^T d(P1)^T)^{-1}. \quad (18)$$

$$P^{old}(c | x_n) = \frac{\exp\left(-0.5 \left|\frac{x_n - B^{old} y_c - t^{old}}{b}\right|\right)}{\sum_{c=1}^{M} \exp\left(-0.5 \left|\frac{x_n - B^{old} y_c - t^{old}}{b}\right|\right)}. \quad (19)$$

## IV. RESULTS

Here we test our method against CPD and LM_ICP on synthetically generated 2D fish rigid with respect to 1) noise Domains in the point sets and 2) number of outliers in constant and moving data set. Fig.2 shows the applied method on 2D fish data set. Blue dots show the constant point set and the red ones are the transformed point set. The Green dots shows the Outliers. In order to test our method, we produced 8 random rotations and 20 random Gaussian noises with the length of 60. In each test we have used one of the produced noises for different lengths and Domains and also one of the Produced rotations. And then we repeat the test for different Gaussian Noise parameters used in both our algorithm and CPD. Over all respectively 9*6400 and 4*14400 different tests has been exerted and the average value for each parameters has shown in Fig4 and Fig5. The Fig4 and Fig5 respectively shows the average amount of Accuracy, MSE and Maximum Iteration for 4 different noise counts in all of the tests and 9 Different Noise Domains in all of the tests as well.

## V. DISCUSSION AND CONCLUSION

. Our method is based on Laplacian distribution mixture model. We suppose that the topology of our data sets follows a Laplacian distribution, so we put appropriate Laplacian distribution on all points in the moving data set and tries to find the most optimum transformation that register the centroids of this distributions into the constant data set. We can see that in all of the aspects from Accuracy of point correspondence detection and mean square error to Maximum Iteration needed for each test, our algorithm has shown remarkable results compare to CPD.

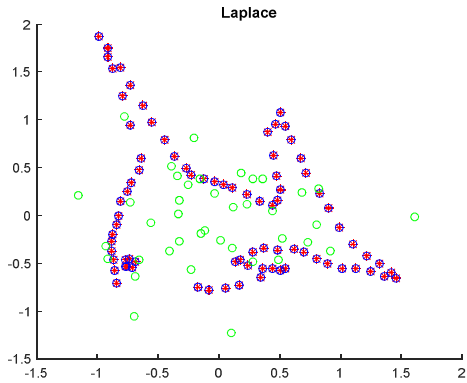

Fig. 2 Rigid Registration of 2D fish data set

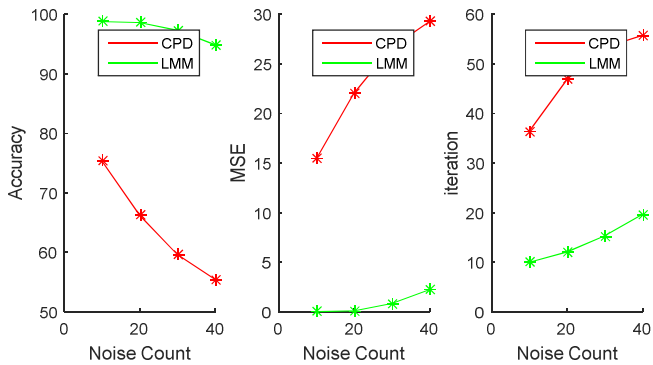

Fig.3 comparison of two methods for different number of noises

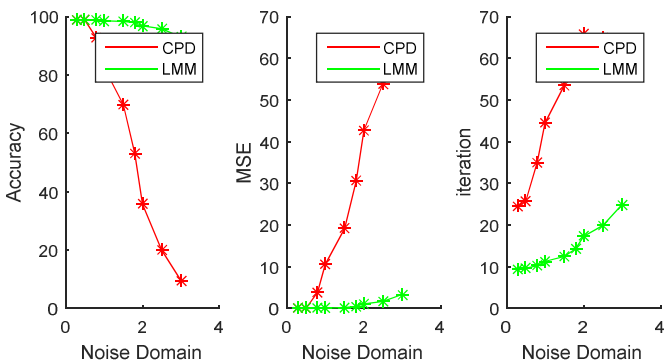

Fig. 4 comparison of two methods for different Noise Domains